\def\year{2021}\relax
\documentclass[letterpaper]{article} 
\usepackage{aaai21}  
\usepackage{amsmath,amssymb,amsfonts}
\usepackage{times}  
\usepackage{helvet} 
\usepackage{courier}  
\usepackage[hyphens]{url}  
\usepackage{graphicx} 
\usepackage[switch]{lineno}
\usepackage{natbib}  
\usepackage{caption} 
\title{Many-to-One Distribution Learning and $K$-Nearest Neighbor Smoothing for Thoracic Disease Identification}

\author{
    Yi Zhou\textsuperscript{\rm 1}\thanks{Corresponding author: Yi Zhou},
    Lei Huang\textsuperscript{\rm 2},
    Tianfei Zhou\textsuperscript{\rm 3},
    Ling Shao\textsuperscript{\rm 4}
    \\
}
\affiliations{
    \textsuperscript{\rm 1}School of Computer Science and Engineering, Southeast University, Nanjing, China\\
    \textsuperscript{\rm 2}State Key Laboratory of Software Development Environment, Beihang University, Beijing, China\\
    \textsuperscript{\rm 3}Computer Vision Laboratory, ETH Zurich, Switzerland\\
    \textsuperscript{\rm 4}Inception Institute of Artificial Intelligence, Abu Dhabi, UAE\\
    yi.zhou@ieee.org

}

\begin{document}

\maketitle

\begin{abstract}
Chest X-rays are an important and accessible clinical imaging tool for the detection of many thoracic diseases. Over the past decade, deep learning, with a focus on the convolutional neural network (CNN), has become the most powerful computer-aided diagnosis technology for improving disease identification performance. However, training an effective and robust deep CNN usually requires a large amount of data with high annotation quality. For chest X-ray imaging, annotating large-scale data requires professional domain knowledge and is time-consuming. Thus, existing public chest X-ray datasets usually adopt language pattern based methods to automatically mine labels from reports. However, this results in label uncertainty and inconsistency. In this paper, we propose many-to-one distribution learning (MODL) and $K$-nearest neighbor smoothing (KNNS) methods from two perspectives to improve a single model's disease identification performance, rather than focusing on an ensemble of models. MODL integrates multiple models to obtain a soft label distribution for optimizing the single target model, which can reduce the effects of original label uncertainty. Moreover, KNNS aims to enhance the robustness of the target model to provide consistent predictions on images with similar medical findings. Extensive experiments on the public NIH Chest X-ray and CheXpert datasets show that our model achieves consistent improvements over the state-of-the-art methods.

\end{abstract}

\section{Introduction}

Chest X-rays are one of the most common radiology exams. Hundreds of millions of such images are acquired in hospitals and clinics all over the world, for identifying a wide range of diseases. Advances in deep learning present a powerful means of developing automated systems to help radiologists interpret chest X-ray images. Recent efforts have shown promise in improving the identification of different chest lesions \cite{rajpurkar2017chexnet,allaouzi2019novel}, localization of their corresponding positions and sizes \cite{li2018thoracic,liu2019align}, and generation of human-readable diagnostic reports \cite{jing2017automatic,li2018hybrid,zhang2020radiology}. However, there exist many drawbacks to chest X-rays, such as the lack of 3D information, inconsistent visual characteristics of various lesions, and label uncertainty, which still make automated chest X-ray diagnosis a very challenging task.

\begin{figure}[t]
\begin{center}
\includegraphics[width=1.0\linewidth]{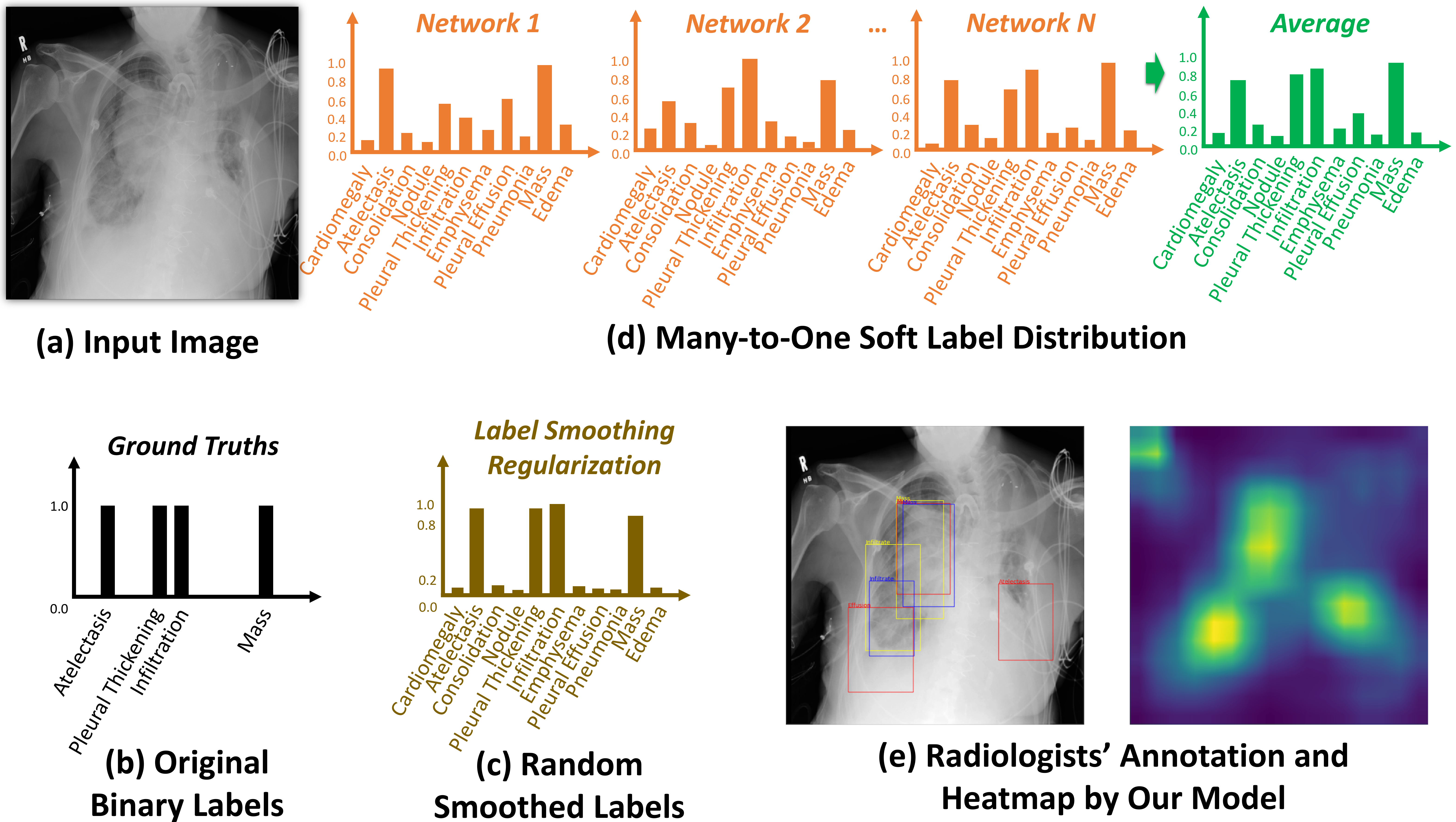}
\end{center}
   \caption{(a) An input image. (b) The ground-truth binary labels automatically extracted from reports. (c) Random smoothed labels proposed by \cite{pham2019interpreting}. (d) Our many-to-one soft label distribution. (e) The bounding boxes of three different colors are annotated by three radiologists, separately, in the left image. The right image is the heat map, showing abnormal regions, obtained by our method.}
\label{fig:motivation}
\end{figure}

For chest X-ray images, large anonymized public datasets are available to researchers, and have facilitated the development of deep learning models. However, providing accurate clinical labels for the very large image sets required for training deep models is difficult and time-consuming. Most datasets \cite{wang2017chestx,irvin2019chexpert} have applied rule-based natural language processing (NLP) to radiology reports to automatically mine the labels, but this often leads to inconsistencies and noise for training. Although previous efforts have tried to address this problem, for example, by converting all the uncertain labels into positive or negative \cite{irvin2019chexpert}, or applying label smoothing regularization \cite{pham2019interpreting}, these methods are not quite effective and only bring a slight improvement. Moreover, most efforts focus on model ensembles when pursuing a further increase of accuracy, rather than enhancing a single model. Combining multiple models, however, will largely increase the computational costs and complexities. Therefore, in this paper, we aim to address these two challenges by exploring label distributions to reduce the effects of uncertainty and noise, and enhancing a single model's ability rather than using a model ensemble. Figure~\ref{fig:motivation} illustrates our motivation and solutions.

To train deep learning models for identifying thoracic diseases, multi-label supervision indicating different lesions is usually adopted. The optimization of such models is treated as an image classification problem that builds a mapping from the instances to the labels via a Softmax cross-entropy or binary cross-entropy loss for each label. However, binary labels (positive or negative) cannot usually describe in detail the severity of a certain disease in an image. As shown in Figure~\ref{fig:motivation} (e), even radiologists are unable to always provide consistent annotations. With the label uncertainty and error further introduced by automatic label extraction, we believe using soft labels of probabilities will contribute to more reasonable training. Thus, in addition to multi-label classification, we propose to optimize the Kullback-Leibler divergence between the ground-truth soft label distribution and the predicted one. Moreover, to obtain a single model with competitive performance to model ensembles, we compute $K$ nearest neighbors using the soft label distributions integrated from multiple models, and then constrain a single target model to provide similar predictions to these scans, which can enhance its robustness. Our main contributions are highlighted as follows:

\textbf{1.} A many-to-one distribution learning (MODL) method is proposed to exploit multiple reference models and integrate their predictions as a soft label distribution to optimize the target network, which can reduce the effects of label uncertainty and ambiguity.

\textbf{2.} To enhance the robustness of a target model for providing consistent predictions on images with similar medical findings, we introduce a $K$-nearest neighbor smoothing (KNNS) module as an auxiliary loss for optimization. The local similarity score is adopted to enhance the learning of the neighbor smoothing.

\textbf{3.} Extensive experiments are conducted to demonstrate the effectiveness of the proposed MODL and KNNS. Consistent improvements by our method are achieved on both the NIH Chest X-ray and CheXpert datasets for thoracic disease identification, compared to state-of-the-art models. Another big advantage of our method is that our single target model does not require additional computational resources during the testing phase.

\section{Related Work}

\subsection{Thoracic Disease Diagnosis}
Recently, deep convolutional neural networks (CNNs) have been extensively investigated for thoracic disease diagnosis \cite{litjens2017survey,shin2016deep,zhou2021contrast,fan2020inf}. The common tasks can be categorized into multi-disease identification \cite{kumar2018boosted,rajpurkar2017chexnet,zhou2018weakly,baltruschat2019comparison}, localization \cite{pesce2019learning,yao2018weakly,liu2019align}, and automatic report generation \cite{wang2018tienet,jing2017automatic,li2018hybrid}. To facilitate research into thoracic disease diagnosis, various public chest X-ray datasets \cite{wang2017chestx,irvin2019chexpert,johnson2019mimic} have been released. Several single- or multi-CNN models \cite{rajpurkar2017chexnet} have been developed for disease classification, while weakly supervised mechanisms using limited bounding-box annotations are typically used for localization. Li \textit{et al.} \cite{li2018thoracic} proposed a patch slicing layer to resize CNN features by max-pooling or bilinear interpolation and applied a fully convolutional recognition network on these regions, to improve identification and localization, simultaneously. Moreover, a contrast induced attention network was introduced by \cite{liu2019align} to exploit the highly structured property of chest X-ray images for localizing diseases. In addition, the automatic generation of medical image reports has also been investigated; for example, \cite{jing2017automatic} exploited a co-attention mechanism to identify abnormal regions and used a hierarchical LSTM to generate sentences. However, while several previous works have mentioned issues with label uncertainty and inconsistency of annotation, but none have provided an effective solution.

\subsection{Reducing Label Inconsistency and Ambiguity}
Label inconsistency, uncertainty, and ambiguity are common problems that affect the training of machine learning models in numerous areas. Several strategies, such as confident learning (CL) \cite{northcutt2019confident}, label distribution learning (LDL) \cite{geng2016label,gao2017deep,chen2020label}, label enhancement (LE) \cite{xu2019partial,xu2018label}, and active learning (AL) \cite{wu2018active}, have been proposed in recent years to mitigate these problems. CL is based on the principles of pruning noisy data, counting to estimate noise, and ranking examples to train with confidence. It can directly estimate the joint distribution of noisy and true labels, and find the label errors. LDL treats the labels representing the degree to which each label describes the instance. It is a more general learning framework of which single-label and multi-label classification are special cases. LDL has been adopted in many applications, including facial landmark detection \cite{su2019soft} and age estimation \cite{hou2017semi}. LE methods aim to solve the unavailability of label distributions. For example, a partial label learning method was presented by \cite{xu2019partial}, in which label distributions are recovered by leveraging the topological information of the feature space. Moreover, AL adopts a small set of clean data, and decreases the noise of the training set by re-labeling other uncertain samples.

\begin{figure*}[t]
\begin{center}
\includegraphics[width=1.0\linewidth]{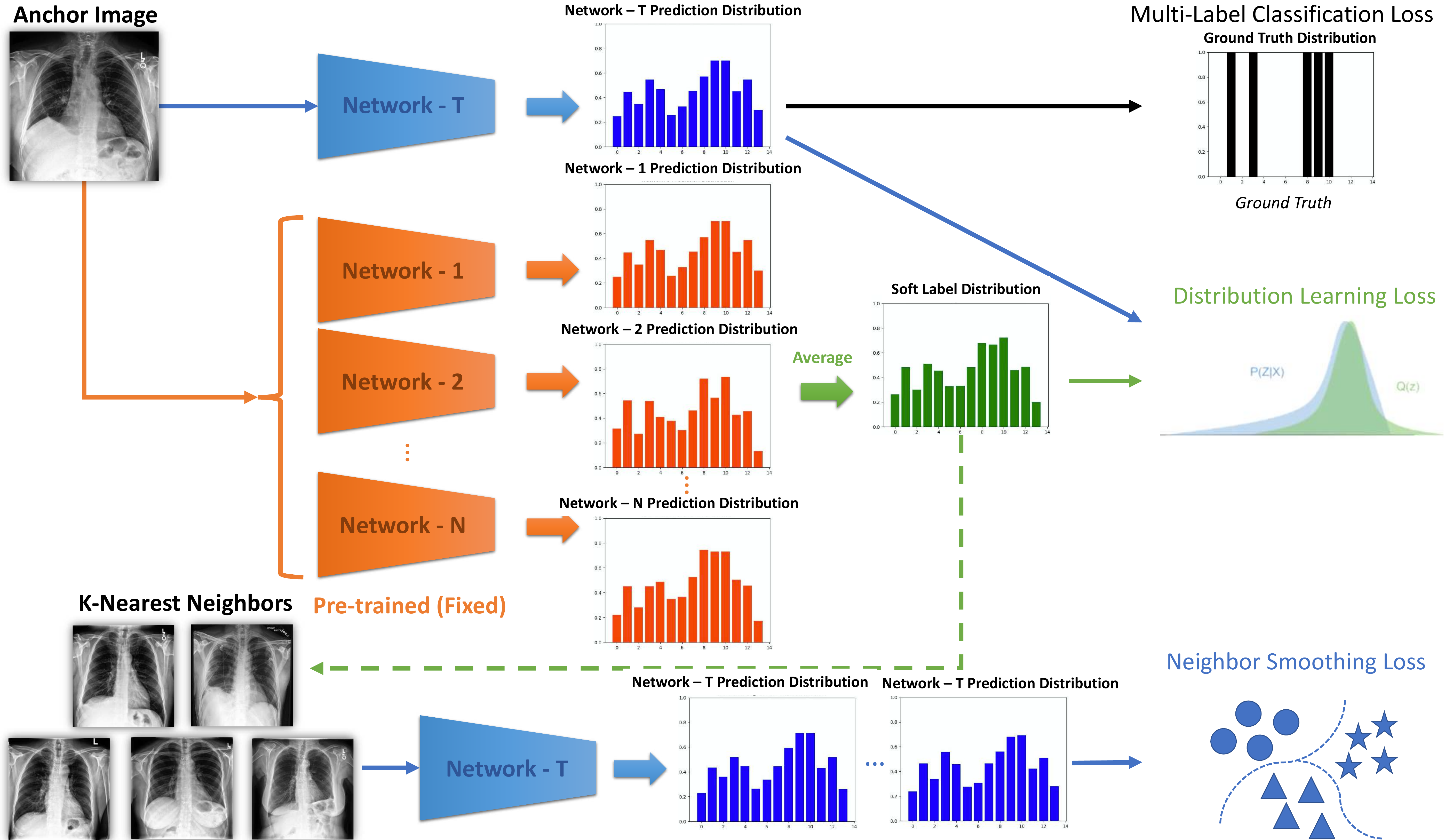}
\end{center}
   \caption{Overview of the proposed method. Network-T is the single target network, which we aim to optimize. $N$ reference networks are pre-trained to compute soft label distributions for helping to optimize Network-T with a distribution learning loss, which can reduce the uncertainty and ambiguity of the original binary labels. Moreover, $K$ nearest neighbors are computed based on the soft label distribution, and then used to constrain the Network-T to provide consistent predictions using an auxiliary neighbor smoothing loss.}
\label{fig:system}
\end{figure*}

\section{Proposed Methods}

\subsection{Problem Formulation}
First of all, the main notations used in this paper are defined as follows. Given the $i$-th input X-ray image $x_i$ and its corresponding binary ground truth labels $y^j_i$, where $j$ denotes the $j$-th disease label, we aim to optimize a single target network $f_T$ for thoracic disease identification. Due to the uncertainty of the original binary labels, we train multiple reference models $f_n, n \in \{1, 2, ..., N \}$, where $N$ is the number of networks, and integrate their pseudo probability predictions as the input image's soft label distribution $l_i = (l_{x_i}^{y_1}, l_{x_i}^{y_2}, ..., l_{x_i}^{y_C})$, where $C$ is the number of label classes. Thus, in addition to the standard multi-label classification loss $\mathcal{L}_{cls}$, we propose a label distribution learning loss $\mathcal{L}_{distri}$ to better optimize the target $f_T$ for reducing the effects of the label uncertainty. Moreover, we indicate the input image as an anchor, and compute its $K$ nearest neighbors $x_i^k, k \in \{1, 2, ..., K\}$ based on the averaged soft label predictions. An auxiliary neighbor smoothing loss $\mathcal{L}_{neigh}$ is introduced to enhance the robustness of the target $f_T$. Therefore, the target model can be optimized by minimizing the overall loss function as:

\begin{align}
\label{eq:e1}
& \mathcal{L} = \sum_{i, j} [\mathcal{L}_{cls}(y_i^j, f_T(x_i | \theta) ) + \lambda \mathcal{L}_{distri}(l_i, f_T(x_i | \theta) ) \\
& \nonumber + \gamma \sum_{k} \mathcal{L}_{neigh}(f_T(x_i | \theta), f_T(x_i^k | \theta) ) ] \text{,}
\end{align}
where $\theta$ is the parameter vector, and $\lambda$ and $\gamma$ balance the weights of different losses. An overview of our proposed method is illustrated in Figure~\ref{fig:system}.

\subsection{Basic Multi-Label Classification and Label Smoothing Regularization}
The basic thoracic disease identification framework used in most previous works \cite{wang2017chestx,rajpurkar2017chexnet}, usually adopts a classic deep CNN (i.e. backbone) and optimizes it with a binary cross-entropy (BCE) loss for multi-label classification. DenseNet-121 \cite{huang2017densely} is the most commonly used backbone for this task, since it improves the flow of information and gradients through the network, making the optimization of deep networks tractable. Therefore, we select DenseNet-121 as our target network $f_T$ for final evaluation, and use the BCE loss as $\mathcal{L}_{cls}$ for training.

To deal with the label uncertainty caused by the ground-truth labels not being annotated by human experts for images provided in the training dataset, several preliminary studies have been conducted. For example, for the large-scale public CheXpert \cite{irvin2019chexpert} dataset, uncertain labels are often extracted due to the unavoidable ambiguities in radiology reports. These labels can all be $ignored$ (U-Ignore), converted to $positive$ labels (U-Ones), or converted to $negative$ labels (U-Zeros), proposed in \cite{irvin2019chexpert}. Moreover, in \cite{pham2019interpreting}, label smoothing regularization (LSR) was applied to better handle these uncertain labels, with the main goal of preventing the model from training with overconfident labels that might contain misclassified data. Specifically, in the U-Ones setting, the uncertainty labels are mapped to a random number close to one, subject to a uniform distribution $\mathcal{U}(a, b)$, where $a$ and $b$ are the hyper-parameters. In contrast, in the U-Zeros setting, a similar process is used to convert the uncertainty labels to a random number close to zero. Although these solutions provide some improvements, they are based on too much on brute-force to address the mislabeled data reasonably.

\subsection{Many-to-One Distribution Learning}
Incorporating the opinions of multiple radiologists is usually more reliable than only adopting a single diagnosis. Similarly, although the target network $f_T$ based on the single backbone DenseNet-121 can achieve satisfactory results, model ensemble of several network architectures usually further makes a big improvement on the mean accuracy. According to many observations from extensive experiments, different network architectures, and even different epoch checkpoints for one network architecture, provide inconsistent predictions and have their own strengths on different diseases. Thus, integrating probability predictions from multiple models can reduce the effects of the uncertainty and ambiguity from the original binary labels. In this paper, we propose a many-to-one distribution learning method to optimize a single target network by taking the essence and discarding the dregs of multiple networks, which can also avoid the extra computational resources introduced by a model ensemble in the testing phase.

To select a strong set of reference models, we follow \cite{pham2019interpreting} and adopt six state-of-the-art CNNs ($N = 6$), including DenseNet-121, DenseNet-169, DenseNet-201 \cite{huang2017densely}, Inception-ResNet-v2 \cite{szegedy2017inception}, Xception \cite{chollet2017xception}, and NASNetLarge \cite{zoph2018learning}. Once the training is separately done for each network, the probability predictions of all the trained networks are simply averaged, and can then be viewed as the soft label distribution. By getting an average, those evident lesions with high probabilities predicted by most of the reference networks still get positive-toward values, and healthy scans still get negative-toward values. However, those lesions getting inconsistent predictions which have ambiguities will be smoothed to reduce the effects of uncertainty. Our label distribution smoothing tends to be a major voting manner, which is more reasonable than the random LSR method. Therefore, this soft label distribution for each image should be closer to the correct labels to indicate the ground truths. We use the Kullback-Leibler (KL) divergence for the distribution learning loss $\mathcal{L}_{distri}$ and minimize it to optimize the target network $f_T$:

\begin{align}
\label{eq:e2}
& \mathcal{L}_{distri} = KL\{l_i || f_T(x_i | \theta)\} = \sum_{i} l_i \log(\frac{l_i}{f_T(x_i | \theta)}) \text{.}
\end{align}

Please note that all the reference networks $f_n$ are pre-trained in advance. Their network parameters are fixed when training the target network $f_T$. Thus, the soft label distribution $l_i$ is fixed to compute the KL divergence as the relative entropy of $l_i$ with respect to $f_T(x_i | \theta)$.

\subsection{$K$-Nearest Neighbor Smoothing}
Our second important concern is how to make the target network as robust and competitive as the model ensemble. According to the smoothness assumption \cite{chapelle2009semi}, we assume that if two samples are similar in the averaged label distribution space, they should also be similar in the label space predicted by the target model. Therefore, before training the target network, the $K$ nearest neighbors of an image (marked as the anchor image) are pre-computed based on the fixed soft label distribution.

To make the disease predictions of the anchor image $f_T(x_i | \theta)$ as close as the predictions of its corresponding $k$-th neighbor $f_T(x_i^k | \theta)$, we generate a pool that contains $K$ neighbor images of each anchor during the training, for enhancing local similarity. Let $k$ denote a sample in the pool. Inspired by \cite{xu2018label}, we specify the local similarity $s_i^k$ which is defined as:

\begin{align}
\label{eq:e3}
& s_i^k = 
\begin{cases}
\mathrm{exp}(-\frac{||l_i - l_i^k||^2}{2 \sigma^2}) & \mbox{if $k \in \{ 1, ..., K \}$} \\
0 & \mbox{otherwise} \text{,}
\end{cases}
\end{align}
where $\sigma$ denotes the width parameter for similarity calculation, which is fixed as 1 in our settings. Thus, the more similar $l_i$ and $l_i^k$ are, the higher $s_i^k$ will be, which means $f_T(x_i | \theta)$ and $f_T(x_i^k | \theta)$ should be closer. We propose a neighbor smoothing loss $\mathcal{L}_{neigh}$ as follows:

\begin{align}
\label{eq:e4}
& \mathcal{L}_{neigh} = \sum_{i, k} s_i^k f_T(x_i^k | \theta) \log \frac{f_T(x_i^k | \theta)}{f_T(x_i | \theta)} \text{.}
\end{align}
By optimizing $\mathcal{L}_{neigh}$, the target model $f_T$ is constrained to give similar predictions of neighboring images which have similar abnormal findings. This enhances the robustness of $f_T$ for more consistent identification performance.

\begin{table*}[t]
\centering
\resizebox{1.0\textwidth}{!}{
\begin{tabular}{c|cccccccccccccc|c}
\hline
Methods & Atel. & Card. & Effu. & Infi. & Mass & Nod. & Pne.1 & Pne.2 & Cos. & Ede. & Emp. & Fibr. & P.T. & Hern & Mean \\ \hline
\cite{wang2017chestx} & 70.03 & 81.00 & 75.85 & 66.14 & 69.33 & 66.87 & 65.80 & 79.93 & 70.32 & 80.52 & 83.30 & 78.59 & 68.35 & 87.17 & 74.51  \\
\cite{yao2018weakly} & 73.30 & 85.80 & 80.60 & 67.50 & 72.70 & 77.80 & 69.00 & 80.50 & 71.70 & 80.60 & 84.20 & 75.70 & 72.40 & 82.40 & 76.73  \\
\cite{yao2017learning} & 77.20 & 90.40 & 85.90 & 69.50 & 79.20 & 71.70 & 71.30 & 84.10 & 78.80 & 88.20 & 82.90 & 76.70 & 76.50 & 91.40 & 80.30  \\
\cite{li2018thoracic} & 80.00 & 81.00 & 87.00 & 70.00 & 83.00 & 75.00 & 67.00 & 87.00 & 80.00 & 88.00 & 91.00 & 78.00 & 79.00 & 77.00 & 80.21 \\
\cite{rajpurkar2017chexnet} & 80.94 & \textbf{92.48} & 86.38 & \textbf{73.45} & 86.76 & 78.02 & 76.80 & 88.87 & 79.01 & 88.78 & 93.71 & 80.47 & 80.62 & 91.64 & 84.13 \\ \hline
Baseline (\textbf{B}) & 82.42 & 89.80 & 89.68 & 68.46 & 86.84 & 80.34 & 75.31 & 85.34 & 80.37 & 91.62 & 90.82 & 84.60 & 80.64 & 93.97 & 84.30  \\
\textbf{B}+MODL & 83.91 & 91.66 & 91.58 & 72.84 & 88.68 & 81.65 & 76.73 & 89.32 & 82.19 & 93.72 & 93.83 & 86.51 & 82.92 & 95.69 & 86.52   \\
\textbf{B}+KNNS & 84.44 & 91.73 & 91.62 & 72.73 & 88.99 & 82.12 & 76.97 & 89.52 & 82.11 & 93.75 & 93.81 & 86.78 & 83.74 & \textbf{95.77} & 86.72   \\
\textbf{B}+MODL+KNNS & \textbf{85.20} & 92.24 & \textbf{92.09} & 73.01 & \textbf{89.74} & \textbf{82.94} & \textbf{78.30} & \textbf{90.46} & \textbf{83.04} & \textbf{94.42} & \textbf{94.32} & \textbf{87.64} & \textbf{84.26} & 95.28 & \textbf{87.35}   \\ \hline
\end{tabular}
}
\caption{Identification results using AUC (\%) metric on the NIH Chest X-ray dataset.}
\label{table:NIH_classification_comparison}
\end{table*}

\begin{table*}[t]
\centering
\resizebox{0.9\textwidth}{!}{
\begin{tabular}{c|ccccc|c}
\hline
Methods & Atelectasis & Cardiomegaly & Consolidation & Edema & Pleural Effusion & Mean \\ \hline
\cite{allaouzi2019novel} LP & 72.0 & 87.0 & 77.0 & 87.0 & 90.0 & 82.6 \\
\cite{allaouzi2019novel} BR & 72.0 & 88.0 & 77.0 & 87.0 & 90.0 & 82.8 \\
\cite{allaouzi2019novel} CC & 70.0 & 87.0 & 74.0 & 86.0 & 90.0 & 81.4 \\
\cite{irvin2019chexpert} U-Zeros * & 81.1 & 84.0 & 93.2 & 92.9 & 93.1 & 88.8 \\
\cite{irvin2019chexpert} U-Ones * & 85.8 & 83.2 & 89.9 & 94.1 & 93.4 & 89.3  \\
\cite{pham2019interpreting} U-Zeros+CT+LSR & 80.6 & 83.3 & 92.9 & 93.3 & 92.1 & 88.4 \\
\cite{pham2019interpreting} U-Ones+CT+LSR & 82.5 & 85.5 & 93.7 & 93.0 & 92.3 & 89.4 \\ \hline
Baseline (\textbf{B}) & 87.29 & 84.47 & 92.09  & 91.66  & 91.34  & 89.37  \\ 
\textbf{B}+MODL &  87.64 &  87.84 & 93.34  & 94.58  & 93.35  & 91.35  \\ 
\textbf{B}+KNNS &  89.15  &  88.24  &  92.40  &  94.92  &  92.67  &  91.48  \\ 
\textbf{B}+MODL+KNNS &  \textbf{89.46} &  \textbf{89.67} & \textbf{93.80}  & \textbf{95.87}  & \textbf{94.35}  & \textbf{92.63}  \\  \hline
\end{tabular}
}
\caption{Identification results using AUC (\%) metric on the CheXpert dataset. * indicates that the result is obtained by the ensemble of 30 checkpoints.}
\label{table:chexpert_classification_comparison}
\end{table*}

\subsection{Implementation Details}
The training procedure of our proposed method consists of two steps. In the first step, all the reference models based on different backbones are pre-trained using the basic multi-label classification loss $\mathcal{L}_{cls}$. All the images are resized into a standard size $256 \times 256$ and randomly cropped to $224 \times 224$ patches as inputs. The Adam optimizer is adopted with an initial learning rate of 0.001 and default parameters $\beta_1 = 0.9, \beta_2 = 0.999$. The network parameters are initialized with a model pre-trained on ImageNet \cite{deng2009imagenet}. The mini-batch size is set to 32 for training over 10 epochs and the learning rate is reduced by a factor of 3 after every two epochs. Once all the reference models are well trained, the soft label distribution of each image is computed and used for training in the second step.

In the second step of the training phase, in addition to $\mathcal{L}_{cls}$, we add the proposed MODL loss $\mathcal{L}_{distri}$ and KNNS loss $\mathcal{L}_{neigh}$ for optimizing the single target model. For each image in a training batch (we marked it as an anchor), its $K$ neighbor images and the corresponding local similarity scores $s_i^k$ are stored in a pool for optimizing $\mathcal{L}_{neigh}$. The different loss weights $\lambda$ and $\gamma$ in Eq.~\ref{eq:e1} are both set to 0.1 in the experiments. All other training settings in the second step are the same as Step 1. Once the training is completed, only the target model $f_T$ is deployed in the testing phase, without any additional computation consumption introduced by model ensemble.

\section{Experiments and Results}

\subsection{Datasets}
\subsubsection{NIH Chest X-ray \cite{wang2017chestx}} is comprised of 112,120 X-ray images with disease labels from 30,805 unique patients, 51,708 of which contain one or more pathologies, while the remaining images have no findings. The 14 different labels are extracted using NLP to text-mine disease classifications from the associated radiology reports, but more than 10\% of the annotations are incorrect. The dataset split setting of experiments for our model follows \cite{rajpurkar2017chexnet}.

\subsubsection{CheXpert \cite{irvin2019chexpert}} consists of 224,316 chest radiographs of 65,240 patients. The disease labels of the training set are labeled for the presence of 14 observations as positive, negative, uncertain, or unmentioned, which also contain inconsistency and ambiguity. However, CheXpert releases a separate validation set of 200 studies, which is annotated by three board-certified radiologists for five diseases: atelectasis, cardiomegaly, consolidation, edema, and pleural effusion. The evaluation on this set is more convincing.

\begin{figure*}[t]
\begin{center}
\includegraphics[width=1.0\linewidth]{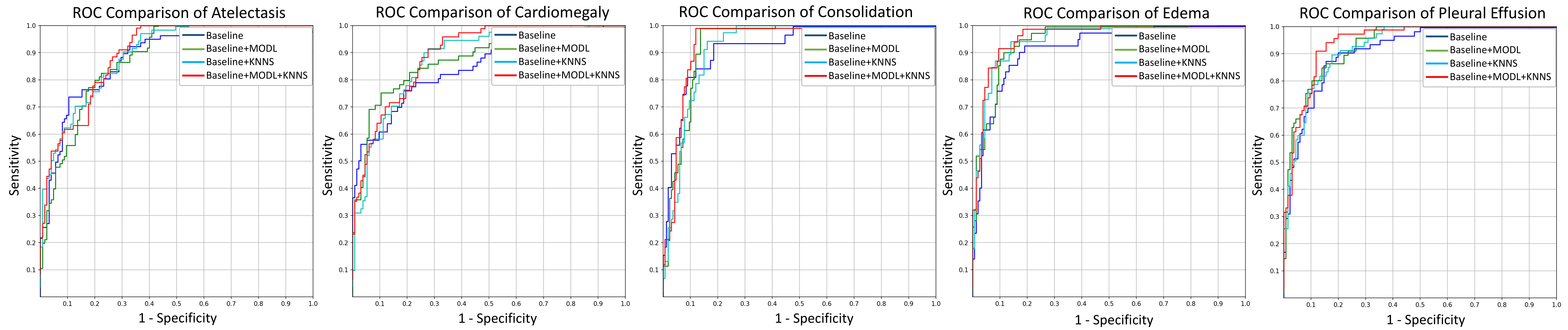}
\end{center}
   \caption{ROC curves of different methods for the five pathologies on the CheXpert dataset.}
\label{fig:roc}
\end{figure*}

\subsection{Comparison with State-of-the-Arts}

To evaluate the multi-disease identification task, the area-under-the-curve (AUC) of the receiver operating characteristic (ROC) is usually adopted. Table~\ref{table:NIH_classification_comparison} provides the identification results on the NIH Chest X-ray dataset. Model \cite{wang2017chestx}, using a ResNet-50 backbone and weighted binary cross-entropy loss, obtains a preliminary mean AUC of 74.51\% over 14 diseases. CheXnet \cite{rajpurkar2017chexnet} adopts DenseNet-121 and obtains a large improvement, since DenseNets improve the flow of information and gradients through the network, making the optimization of deep networks tractable. In \cite{yao2017learning}, a long short-term memory (LSTM) architecture is used to model dependencies among labels, but achieves poor results. For most diseases, our method with many-to-one distribution learning and $K$-nearest neighbor smoothing obtains consistent increases in identification performance, compared to the best-performing method \cite{rajpurkar2017chexnet}. The only exception is on cardiomegaly and infiltration. The mean AUC is increased by 3.22\%.

The results on CheXpert are compared in Table~\ref{table:chexpert_classification_comparison}. Most methods exploit ensembles of multiple models to pursue higher scores. For example, an ensemble of 30 generated checkpoints of DenseNet-121 \cite{irvin2019chexpert} achieves a mean AUC of 89.3\%. In this subsection, we mainly focus on comparing the results achieved by single models. Binary relevance (BR), label powerset (LP), and classifier chain (CC) are explored in terms of label dependencies in \cite{allaouzi2019novel}, but no satisfactory performance is achieved. Conditional training (CT) and label smoothing regularization (LSR) are used to enhance the performance of a single DenseNet with a small gain achieved. In our implementation, the U-Ones setting is adopted to deal with those uncertainty labels when training the Baseline (\textbf{B}) (based on the backbone DenseNet-121), as well as when training the reference models in the first training stage. Our method \textbf{B}+MODL+KNNS without any ensemble operation increases the mean AUC by 3.26\%.

\subsection{Ablation Studies}

\subsubsection{Effectiveness of MODL}
To evaluate the effectiveness of the proposed many-to-one distribution learning, the soft label distribution obtained is only used to optimize the target network through the auxiliary loss $\mathcal{L}_{distri}$. Clear improvements in mean AUC 2.22\% and 1.98\% are achieved on NIH Chest X-ray and CheXpert, respectively, compared to the baselines. This illustrates that the soft label distribution can represent the ground truths of a scan better than the original automatically extracted binary labels which have uncertainty and ambiguity.

\begin{figure}[t]
\begin{center}
\includegraphics[width=0.8\linewidth]{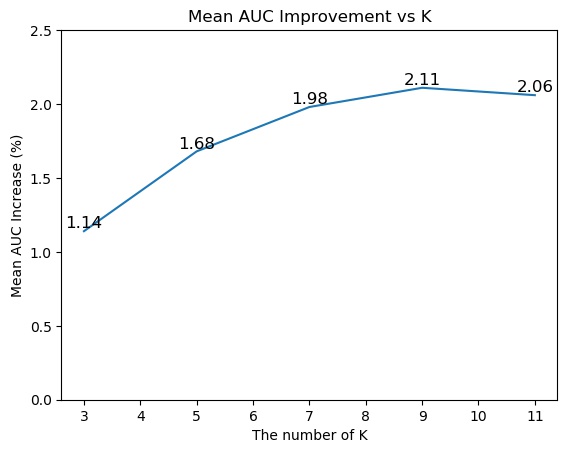}
\end{center}
   \caption{Comparison of the mean AUC increase based on different $K$ in KNNS.}
\label{fig:knns}
\end{figure}

\subsubsection{Effectiveness of KNNS}
The $K$-nearest neighbor smoothing is also separately validated. An anchor image and its $K$ neighbor images are passed forward to the target network, which is updated by the neighbor smoothing loss $\mathcal{L}_{neigh}$. The results show that constraining the model to provide consistent predictions for images with similar medical findings also largely improves the performance. The mean AUC is increased by 2.42\% and 2.11\% on NIH Chest X-ray and CheXpert, respectively. Moreover, combining $\mathcal{L}_{distri}$ and $\mathcal{L}_{neigh}$ can further enhance the target model without requiring any extra computational resources during the testing phase. Figure~\ref{fig:roc} provides the ROC curves of different methods for the five pathologies on the CheXpert validation set.

\subsubsection{Investigation of the Number $K$ of KNNS}
Another concern is to determine the best-performing $K$ of KNNS. We conduct an experiment on the CheXpert dataset, where $K$ is increased from 3 to 11 by a step of 2. The mean AUC improvement in terms of $K$ is shown in Figure~\ref{fig:knns}. We observe that configuring three neighbor images to constrain the model prediction consistency can already obtain a significant increase in mean AUC of 1.14\%. Improvements of 1.68\% and 1.98\% are achieved when we increase $K$ to 5 and 7, respectively. However, with the further growth of $K$, the increase rate becomes smaller and the method requires much higher computational memory during the training phase. Therefore, considering the trade-off between performance and computational costs, we set $K$ as 9 in KNNS.

\begin{table}[t]
\centering
\resizebox{0.4\textwidth}{!}{
\begin{tabular}{c|cc}
\hline
Methods          & Mean AUC & Params. \\ \hline
DenseNet-121 (\textbf{B}) &    89.66      & 7.98 M  \\
\textbf{B}+MODL+KNNS      &    92.63      & 7.98 M  \\
Model Ensemble 1 &    92.19      &     42.13 M    \\
Model Ensemble 2 &     93.96     &     208.33 M    \\ \hline
\end{tabular}
}
\caption{Comparison with model ensembles on the CheXpert dataset. M denotes million.}
\label{table:ensemble}
\end{table}

\begin{figure*}[t]
\begin{center}
\includegraphics[width=1.0\linewidth]{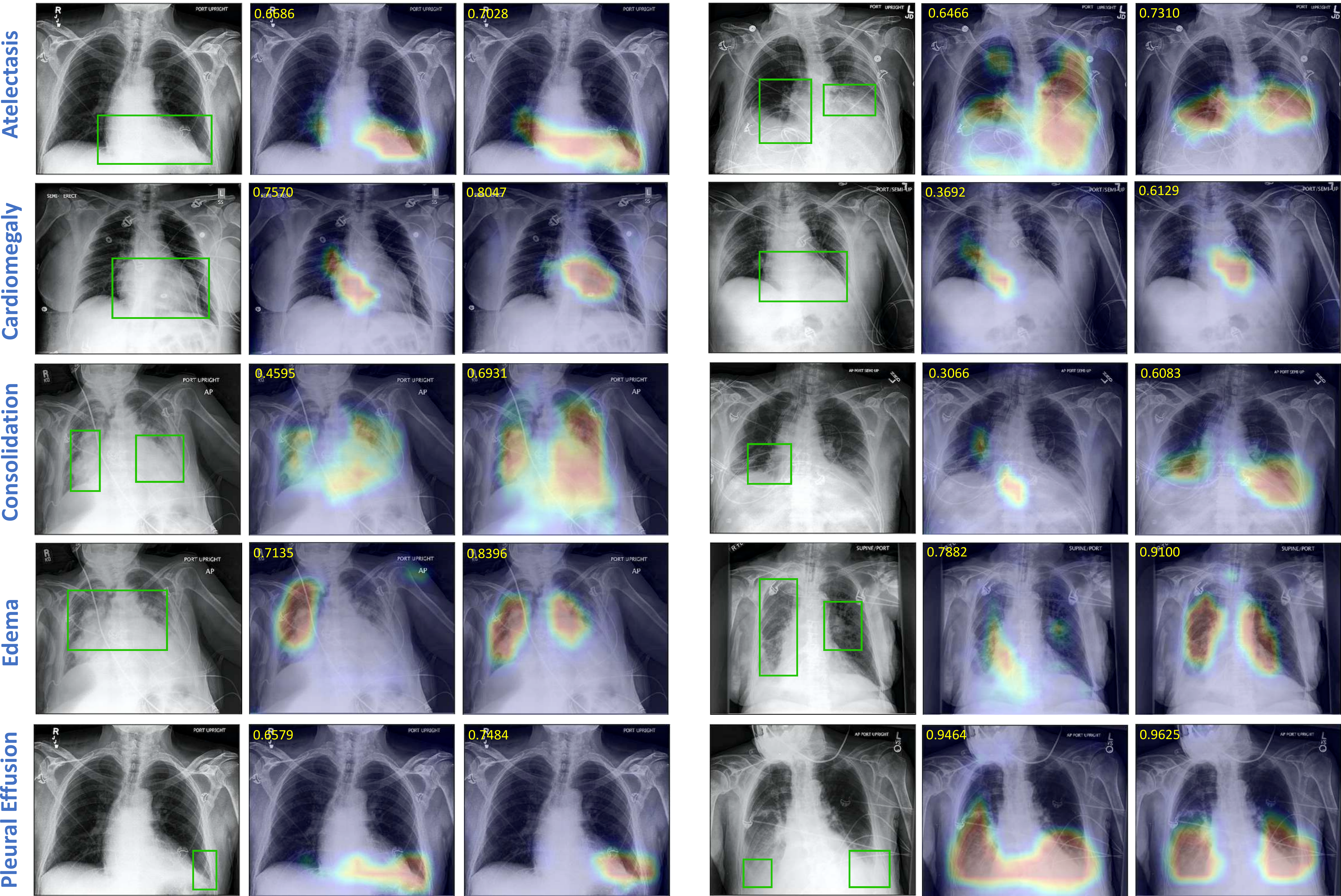}
\end{center}
   \caption{Heatmap visualization on the CheXpert dataset. Each row shows two cases of one disease. For each case, the left is the input image with ground truths. The middle and right are the heatmaps obtained by the Baseline (\textbf{B}) and \textbf{B}+MODL+KNNS, respectively. The value on the top-left of each heatmap is the predicted probability of the corresponding disease.}
\label{fig:demonstration}
\end{figure*}

\subsubsection{Comparison with Model Ensembles}
In addition to the mitigation of label uncertainty and enhancement of model robustness, another big advantage of our method is the zero-increase in network complexity during the testing phase, compared to model ensemble methods. To validate this point, the mean AUC and number of network parameters of some ensemble models are compared in Table~\ref{table:ensemble}. Model Ensemble 1 is a combination of DenseNet-121, DenseNet-169, and DenseNet-201. Model Ensemble 2 is an ensemble of the six reference models used for learning the soft label distribution in our method. As shown from the results, our method \textbf{B}+MODL+KNNS outperforms Model Ensemble 1 by 0.44\% of the mean AUC, with much fewer parameters. Moreover, although Model Ensemble 2 obtains better identification performance than ours, it has 26 times more parameters than our target model, making it too heavy to deploy in practical applications.

\subsection{Qualitative Results}
To better visualize the model performance, we adopt the probabilistic class activation map (PCAM) \cite{ye2020weakly} pooling to predict the lesion heatmaps with only image-level supervision. As shown in Figure~\ref{fig:demonstration}, we visualize the heatmaps of the five pathologies in the CheXpert dataset. All the examples do have the corresponding pathology and are labeled positive. For each case, we compare the heatmaps obtained by the Baseline (\textbf{B}) and \textbf{B}+MODL+KNNS. In addition to the improvements in identification results demonstrated before, we observe that a more satisfactory localization performance can also be achieved. For all the pathologies, the lesion regions predicted by \textbf{B}+MODL+KNNS are more precise in location and size than those predicted by the Baseline (\textbf{B}). For example, the cardiomegaly heatmaps predicted by the Baseline (\textbf{B}) usually have an abnormal region around the location of the mediastinum, while \textbf{B}+MODL+KNNS can better highlight the enlarged heart shape region. To localize the position of pleural effusion, \textbf{B}+MODL+KNNS can better identify the left heart border, costophrenic angle and hemidiaphragm are obscured, and slight blunting of the right costophrenic angle. Moreover, as shown from the predicted probabilities in the top-left of each heatmap, \textbf{B}+MODL+KNNS identifies the positive diseases as true with higher confidence scores, compared to the Baseline (\textbf{B}).

\section{Conclusion}
In this paper, we proposed a many-to-one distribution learning and a $K$-nearest neighbor smoothing method to improve the performance of thoracic disease diagnosis. The methods can reduce the effects of label uncertainty and ambiguity, and also constrain the target model to provide consistent predictions on images with similar medical findings. Extensive experiments demonstrated that our method is effective and achieves significant improvements without requiring any additional computational costs in the testing phase. In our future work, we will explore visual attention methods to learn better representations of thoracic diseases for improvements.

\bibliography{ref} 
\bibliographystyle{aaai}

\end{document}